\pdfoutput=1

\documentclass[11pt]{article}

\usepackage[]{emnlp2021}

\usepackage{times}
\usepackage{latexsym}

\usepackage[T1]{fontenc}

\usepackage[utf8]{inputenc}

\usepackage{microtype}

\title{The Effects of In-Domain Corpus Size on pre-training BERT}

\author{Chris Sanchez$^{a}$, Zheyuan Zhang$^{b}$$^{*}$ \\
    \small $^{a}$Microsoft Corporation, Reston, Virginia 20191\\
        \small $^{b}$Coleridge Initiative \\
        \small $^{*}$Corresponding author: Zheyuan Zhang: jason.zhang@coleridgeinitiatve.org \\
  }

\begin{document}
\maketitle
\begin{abstract}
Many prior language modeling efforts have shown that pre-training on an in-domain corpus can significantly improve performance on downstream domain-specific NLP tasks. However, the difficulties associated with collecting enough in-domain data might discourage researchers from approaching this pre-training task. In this paper, we conducted a series of experiments by pre-training Bidirectional Encoder Representations from Transformers (BERT) with different sizes of biomedical corpora. The results demonstrate that pre-training on a relatively small amount of in-domain data (4GB) with limited training steps, can lead to better performance on downstream domain-specific NLP tasks compared with fine-tuning models pre-trained on general corpora.\footnote{GitHub repo: https://github.com/JasonZhangzy1757/the-effect-of-domain-corpus-size-for-pretraining}
\end{abstract}

\section{Introduction}

Pre-training large neural language models based on Transformers \citep{vaswani2017attention} such as Bidirectional Encoder Representations from Transformers (BERT) \citep{devlin2018bert} and its variants \citep{liu2019roberta, yang2019xlnet, lan2019albert} has proven to be an excellent strategy and achieved state-of-the-art results on many downstream natural language processing (NLP) tasks. Most models focused their pre-training efforts on general domain text. For example, the original BERT model was trained on Wikipedia and the BookCorpus \citep{zhu2015aligning}. Many other following efforts focused on adding additional texts to the pre-training process to create even larger models with the intent of improving model performance \citep{liu2019roberta, raffel2019exploring}. However, recent works have shown that, given the general nature of the corpora these models were pre-trained on, they do not perform as well when introduced to domain-specific corpora as found in categories such as biomedicine, law, or finance, to name a few.   Several efforts have demonstrated that by pre-training on domain-specific corpora, either through a “from scratch” or a continual pre-training approach, these same models achieved much better performance on in-domain, downstream NLP tasks. \citep{beltagy2019scibert, lee2020biobert, huang2019clinicalbert}

The success of these domain-specific BERT models encouraged practitioners to explore the possibilities of pre-training BERT using corpora in their respective domains to get better performing language models and to better tackle their in-domain NLP tasks. One of the biggest challenges presented to NLP practitioners, however, lies in the lack of large, readily available domain-specific corpora; for reference, the original BERT was pre-trained on 3.3 billion tokens, or roughly 20 GB of uncompressed raw text. To get enough in-domain text data for pre-training, practitioners must resort to what is either available “in-house” or available through public or private resources.  Web scraping is one oft-cited method used to gather publicly available documents to increase one’s in-domain training corpora. For example, LEGAL-BERT \citep{chalkidis2020legal} authors scraped publicly available legal text from six different sources, to achieve a total corpus size of 12 GB.  Nevertheless, this data collection process is laborious and time-consuming and could discourage researchers from conducting such experiments for fear of being unable to collect enough data. On the other hand, it would also be a waste of resources if, after all the data is collected, it turns out the data is still not enough for pre-training and the model ends up having poor performance.  

To mitigate these potential situations, we conducted a series of experiments that pre-train the BERT model on different in-domain corpus sizes and evaluate the resulting language model on multiple downstream, in-domain NLP tasks. Because the biomedical field has a rich history of NLP work, and therefore has several readily available datasets for model fine-tuning/testing, we chose to use biomedicine for our example domain.  Using this example, we demonstrate the least amount of in-domain data required to see performance improvements against generally-trained BERT models across a variety of domain-specific tasks.  We believe this work is useful to practitioners who are considering pre-training their own in-domain BERT model from scratch.  They can use our work to inform their own costs-benefits analysis as they consider whether they have enough resources (data/compute/time) and if the potential gains are worth the pre-training undertaking in the first place.

\begin{table}
\centering
\begin{tabular}{p{0.4\linewidth}|p{0.25\linewidth} p{0.2\linewidth}}
\hline
\textbf{Dataset} & \textbf{Task} & \textbf{Metrics} \\
\hline
NCBI-disease & NER & Micro F1 \\
PubMedQA & QA & Accuracy \\
HoC & Classification & Micro F1 \\
\hline
\end{tabular}
\caption{Datasets for Fine-tuning Tasks and Their Evaluation Metrics}
\label{1}
\end{table}

\section{Related work}

Many prior works have been completed on the domain-adaptation of the BERT model. There are, in general, three possible approaches: a) One could fine-tune a BERT model against in-domain datasets, as suggested by the original BERT authors \citep{devlin2018bert}. However, simple fine-tuning usually leads to unsatisfactory performance on specific domains where the vocabularies have different distributions from the general text, such as medical clinical notes, legal notes, or biomedical literature \citep{lee2020biobert, huang2019clinicalbert, chalkidis2020legal}. b) Intuitively, one could pre-train BERT from scratch using domain-specific corpora with a new vocabulary. This method has proven effective in many specific domains with an abundant amount of specialized vocabulary or particular syntax. Even in the domains that are not usually considered distinctly different from the general text, such as Twitter and Yelp, pre-training on domain-specific corpus still helps improve downstream performance \citep{dai2020cost}. The shortcoming, though, is that this approach typically requires a large amount of data and can be expensive in terms of compute resources and the time/labor costs required to pre-train a model from scratch \citep{tai2020exbert}. c) Another recent prevailing method is to pre-train using a mixed-domain, where the pre-trained BERT model is continually pre-trained on in-domain data, starting from a predefined general model checkpoint. This method assumes the general text is still helpful and the goal is to improve model performance without having to pre-train the model from scratch, thereby reducing the required in-domain corpora size. Several works have explored this method either directly using BERT’s original vocabulary \citep{lee2020biobert}, or incorporating a set of new domain vocabulary into the existing model \citep{tai2020exbert}. Though this method has proved to be effective, there are also works pointing out that pre-training entirely from scratch using only in-domain-specific corpora can significantly outperform models using the continual pre-training approach \citep{gu2021domain}.

There are many experiments conducted to compare the performance between these above-mentioned methods \citep{gu2021domain}, the effect of the domain-specific vocabulary as well as the model size \citep{tai2020exbert}. But to our best knowledge, there are yet no prior works focusing on systematically analyzing the effect of the pre-training corpus size itself. 

\begin{table*}
\centering
\begin{tabular}{llllll}
\hline
\textbf{Dataset} & \textbf{BERT} & \textbf{PubMedBERT} & \textbf{4GB} & \textbf{8GB} & \textbf{12GB} \\
\hline
NCBI-disease & 84.3 & 87.8 & 87.7 & 87.9 & 88.0 \\
HoC & 79.0 & 82.3 & 81.1 & 82.5 & 81.4 \\
PubMedQA & 54.4 & 55.8 & 54.9 & 53.4 & 55.2 \\
\hline
\end{tabular}
\caption{\label{citation-guide}
Performance comparison of pre-trained language models. The models are evaluated on the tasks using the same fine-tuning process. All of our experimental models are pre-trained for 67K steps.}
\label{2}
\end{table*}

\section{Data}

\subsection{Pre-training corpus}

The pre-training corpus was collected from the National Institute of Health’s National Library of Medicine’s PubMedCentral via the AWS public registry \citep{sayers2021database}.  PubMed is a free search engine providing access to papers and scholarly articles primarily focused on biomedical and life sciences topics. The data includes PubMed abstracts and PubMed Central full text articles. For our experiment, we collected 16 GB of full text PubMed articles, discarded all foreign language articles, and used only the main body of the text. 

As part of our prepossessing pipeline, we created a set of vocabulary based on the PubMed corpus and the vocabulary size was set at 30,500 to mimic the original BERT. Of note, the overlap between the new tokenized vocabulary and the original BERT’s vocabulary is only 52\%, in essence, showcasing the dramatic difference in biomedical jargon as compared to words used in general corpora. All words were tokenized as lowercase.

\subsection{Fine-tuning tasks}

The best way to evaluate the effects of model pre-training is to compare the resulting model performance on a range of in-domain NLP tasks against the original BERT model. In order to systematically evaluate the performance of the models, we selected a subset of tasks from the Biomedical Language Understanding \& Reasoning Benchmark (BLURB) benchmark, which comprises a comprehensive set of biomedical NLP tasks from publicly available datasets \citep{gu2021domain}.  The tasks we chose for our experiment include Named Entity Recognition (NER), Question Answering (QA), and Document Classification. We list the datasets we used in Table~\ref{1} as well as the detailed description of the tasks below.

\subsubsection{NCBI-disease} 
The Natural Center for Biotechnology Information Disease corpus \citep{dougan2014ncbi} is fully annotated at the mention and concept level to serve as a research resource for the biomedical natural language processing community. It contains 793 PubMed abstracts and 6892 annotated disease mentions linked to 790 unique disease concepts. For this task, we used the train, development, test splits given by the paper authors.

\subsubsection{PubMedQA}
The PubMedQA \citep{jin2019pubmedqa} is a biomedical question answering dataset collected from PubMed abstracts. The task of PubMedQA is to answer research questions with yes/no/maybe using the corresponding abstracts. PubMedQA has 1k expert-annotated, 61.2k unlabeled and 211.3k artificially generated QA instances. This task has many training options. For simplicity, here we used only the labeled data and the train/test splits provided by the authors. The long answers are also available in the data, but we did not involve them in the experiments.

\subsubsection{HoC}
The Hallmarks of Cancer corpus \citep{hanahan2000hallmarks} contains annotations from PubMed abstracts with labels signifying a specific cancer hallmark. There are 37 detailed hallmarks but we only focused on the top-levels, which are 10 groups in total. We have to split the train/test sets on our own, because they are not provided in the paper. Though the original dataset provided sentence level annotation, we follow the common practice and evaluate on the abstract level \citep{du2019ml, zhang2014zhou}.

\begin{table*}
\centering
\begin{tabular}{llllll}
\hline
\textbf{Dataset} & \textbf{BERT} & \textbf{PubMedBERT} & \textbf{3.5K steps} & \textbf{67K steps} & \textbf{130K Steps} \\
\hline
NCBI-disease & 84.3 & 87.8 & 87.0 & 87.7 & 87.0 \\
PubMedQA & 79.0 & 82.3 & 78.6 & 81.1 & 81.2 \\
HoC & 54.4 & 55.8 & 55.2 & 54.9 & 55.2 \\
\hline
\end{tabular}
\caption{\label{citation-guide}
Performance comparison between general BERT baseline, PubMedBERT paper and different steps of our model pre-trained on 4 GB data.}
\label{3}
\end{table*}

\begin{table*}
\centering
\begin{tabular}{p{0.15\linewidth} p{0.1\linewidth} p{0.1\linewidth} p{0.2\linewidth} p{0.1\linewidth}}
\hline
\textbf{Price/hour (8 x V100s)} & \textbf{Hours Trained} & \textbf{Total Cost} & \textbf{Approximate Score Boost} & \textbf{Price per \% Boost} \\
\hline
\$22.33 & 48 & \$1,071.84 & 2\% & \$535 \\
\$22.33 & 96 & \$2,143.68 & 3.25\% & \$659 \\
\hline
\end{tabular}
\caption{The performance improvement and its approximate cost.}
\label{4}
\end{table*}

\subsection{Experiment Setup}
For the pre-training phase, the models were pre-trained only on the Masked Language Modeling learning objective (masked at 15\%), as a review of the literature indicated that (NSP) is not a necessary loss objective to include in pre-training to achieve SOTA results on downstream tasks \citep{liu2019roberta}. In order to observe the effect of corpus size on model performance, we segmented the data into 4 GB-sized chunks, representing approximately 400,000 documents per chunk.  Literature has demonstrated that 20 GB of text data is a good benchmark for pre-training a domain-specialized BERT model, and therefore we experimented with pre-training runs on corpus chunk sizes of 4 GB, 8 GB, and 12 GB respectively.  Each corpus was an additive version of the previous size. For example, the 8GB corpus consisted of the original 4 GB corpus plus an additional 4 GB of raw text.

For the downstream tasks, we fine-tuned an original BERT model(bert-base-uncased) on each task to set a baseline, and took note of the hyperparameters used.  We then conducted the same fine-tuning using the models pre-trained on 4 GB, 8 GB, and 12 GB of PubMed data, using the same hyperparameters from the original BERT model.. We also conducted the fine-tuning on the downstream tasks of our pre-trained models at various pre-training step sizes to showcase the effects of training duration on model performance. Finally, we report the average scores from five runs for each model on each fine-tuning task.

The experiments were conducted on Microsoft Azure virtual machines, specifically, the Standard ND40rs\_v2  model which consists of 8 x NVIDIA V100 GPUs.  At a Batch Size of 112 and a corpus size of 4 GB, it took roughly 48 hours to pre-train a model up to 130,000 steps.

\section{Results}

We present the results of domain-specific pre-training from scratch on biomedical NLP applications. In this paper, we compare our experimental results against the original BERT model which was trained on a general corpus of 20 GB data, as well as PubMedBERT, which is a mature domain-specialized model also pre-trained using a PubMed corpus, as a reference. The results are demonstrated in two aspects.

Table~\ref{2} summarizes the results of our in-domain models pre-trained on three different sizes of corpus for 67K steps against the general BERT baseline. We can observe, not surprisingly, that the general trend indicates that the larger the corpus size and the longer the model is pre-trained, the better the results tend to be. However, we can also see that the improvement between the 8GB and 12 GB models is less obvious than that between the 4GB and 8GB models. The performance of the 12GB model is fairly close to, in some tasks even slightly better than, the performance of the PubMedBERT model, though the PubMedBERT model was pre-trained on approximately 21 GB of in-domain data. 

It’s worth noting that even though the results of the model trained on 4 GB of data are lower than the models pre-trained on a larger corpus, the performance is just as good, if not better than the general BERT model pre-trained on 20 GB and for 1 million steps. Table~\ref{3} demonstrates the results of the 4GB model after different training steps. We can see from the results that even after one pass through the data (3,500 steps at a batch size of 112), the model is learning how to represent the in-domain language. 

As mentioned previously,  biomedical text is drastically different from general text, and we postulate that the tokenization of in-domain data using an in-domain vocabulary contributes to the improvements in model performance, particularly with sequence classification tasks.  The model is able to learn whole word representations of terms common to the biomedical domain, such as “gastrointestinal”, which would otherwise be broken up into several sub-word tokens if a general vocabulary were used. 

For practitioners we present a simple cost-benefits analysis using our work as an example.  To achieve a roughly 2\% improvement in downstream tasks over general BERT required 48 hours of in-domain pre-training.  Another 48 hours of training led to a further 1-1.5\% jump in performance.  At a run-rate of \$22.33/hour (for 8 x V100s on Azure cloud) the financial cost to achieve the initial performance boost is \$1,071, and if pre-training is run for 96 hours the cost is \$2,143. As shown in Table~\ref{4}, the general trend is the longer you trained, the more you will spend for each percent of improvement. We note here that this analysis is representative of the experiment that we conducted, and can easily be improved upon by simply adding optimization techniques such as Whole Word Masking, faster GPUs, and/or the use of deep learning optimizations such as DeepSpeed \citep{rajbhandari2022deepspeed} or NVIDIA mixed precision  \citep{micikevicius2017mixed}.

\section{Conclusion}

In this paper, we pre-trained the BERT model on different sizes of in-domain corpus and compared the results with the original BERT model. The results demonstrate that even  pre-training on a relatively small amount of in-domain data (4 GB) with limited training steps, can lead to better performance on downstream domain-specific NLP tasks compared with fine-tuning models pre-trained on general corpora. We hope this work could encourages researchers and practitioners who want to pre-train BERT models to solve tasks in specialized domains, but lack access to voluminous stores of raw text data.

\section*{Future Work}
For future work, it would be useful to compare model performance using continual pre-training methods vs.models pre-trained from scratch on comparable corpus sizes, thereby working towards defining a generalized inflection point against which practitioners can weigh their options on which method to use taking into account the text data available, compute resources, time available, and desired model performance.  

Another area that was not explored in our experiment is the effect of vocabulary size on model performance as a function of the corpus size.  We set our vocabulary size at 30,500 to mimic the original BERT model, but as pointed out the original BERT model was trained on 20 GB of data.  We do not know if our models would have benefited from a smaller vocabulary size based on a ratio between the corpus size and vocab size.  

We also readily point out that there are several areas for improvement in our results, starting with using Whole Word Masking as the word masking scheme.  This technique was not used in our experiments but could easily be integrated into the pre-processing pipeline by future practitioners.  We did not experiment with different batch sizes, we simply set the batch size at the maximum memory load our GPUs could handle in order to reduce training time, therefore, there is considerable room for improvement in determining ideal batch size given a set corpus size.  Finally, we would have preferred to test our pre-trained models on a wider variety of downstream tasks so as to determine if our pre-training method is robust across several dimensions.  Due to the time constraints on our project, we were unable to do so, but leave additional fine-tuning on a wider variety of tasks (the Biomedical Language Understanding \& Reasoning Benchmark (BLURB) \citep{gu2021domain} is a great place to start) for future work.  We note that performance gains from in-domain pre-training on some downstream NLP tasks does not necessarily imply that a practitioner will necessarily see gains in other unrelated downstream NLP tasks.  Results on non-tested downstream tasks must be empirically validated. 

\section*{Acknowledgement}
We thank our instructor Prof. Chris Potts and our Course Facilitator Ankit Chadha for taking the time to advise us over the course of this project on matters both great and small. We appreciate their wisdom and candor. We also want to thank Hoifung Poon and Naoto Usuyama from Microsoft Research, as well as Kexin Huang, the clinicalBERT author, who sparked our motivation to attempt this undertaking in the first place.  We are grateful for their advice at the initial stages of our research, which set us on a good trajectory.


\end{document}